\documentclass[10pt,twocolumn,letterpaper]{article}

\usepackage{cvpr}

\usepackage{wrapfig}
\usepackage{algorithm}
\usepackage{algpseudocode}
\usepackage{multirow} 
\usepackage{makecell}
\newcommand{\Normal}[2]{\mathcal{N}\!\left(#1,#2^2\right)}
\newcommand{\Uniform}[2]{\mathcal{U}\!\left(#1,#2\right)}
\usepackage{placeins}
\usepackage{needspace}
\newcommand{\nameCOLOR}[1]{\textcolor{black}{#1}}
\newcommand{\name}{TrajectoryMover\xspace}
\newcommand{\nameMethod}{\nameCOLOR{\mbox{TrajectoryMover}}\xspace}
\newcommand{\nameDataset}{\nameCOLOR{\mbox{TrajectorySynth}}\xspace}

\newcommand{\src}{\text{src}}
\newcommand{\trg}{\text{trg}}
\newcommand{\sourcevideo}{V_{\src}}
\newcommand{\targetvideo}{V_{\trg}}
\newcommand{\controldisp}{I_\text{bb}}
\newcommand{\numf}{F}
\newcommand{\numchannel}{C}
\newcommand{\width}{W}
\newcommand{\height}{H}
\newcommand{\videospace}{\mathbb{R}^{\numf \times \numchannel \times \height \times \width}}
\newcommand{\numflt}{F'}
\newcommand{\numchannellt}{C'}
\newcommand{\widthlt}{W'}
\newcommand{\heightlt}{H'}
\newcommand{\latentspace}{\mathbb{R}^{\numflt \times \numchannellt \times \heightlt \times \widthlt}}

\usepackage{wrapfig}
\usepackage{algorithm}
\usepackage{algpseudocode}
\usepackage{xspace}
\usepackage{tabularx}

\usepackage{chngcntr}

\definecolor{cvprblue}{rgb}{0.21,0.49,0.74}
\usepackage[
    breaklinks,
    colorlinks,
    allcolors=cvprblue
]{hyperref}

\title{TrajectoryMover: Generative Movement of Object Trajectories in Videos}

\author{
Kiran Chhatre$^{1,2}$ \qquad
Hyeonho Jeong$^{2}$ \qquad
Yulia Gryaditskaya$^{2}$ \\
Christopher E. Peters$^{1}$ \qquad
Chun-Hao Paul Huang$^{2}$ \qquad
Paul Guerrero$^{2}$ \\[0.5em]
$^{1}$KTH Royal Institute of Technology \qquad
$^{2}$Adobe Research
}
\begin{document}

\makeatletter
\apptocmd\@maketitle{
    \centering
    \vspace{-1em}

    \includegraphics[width=\textwidth]
    {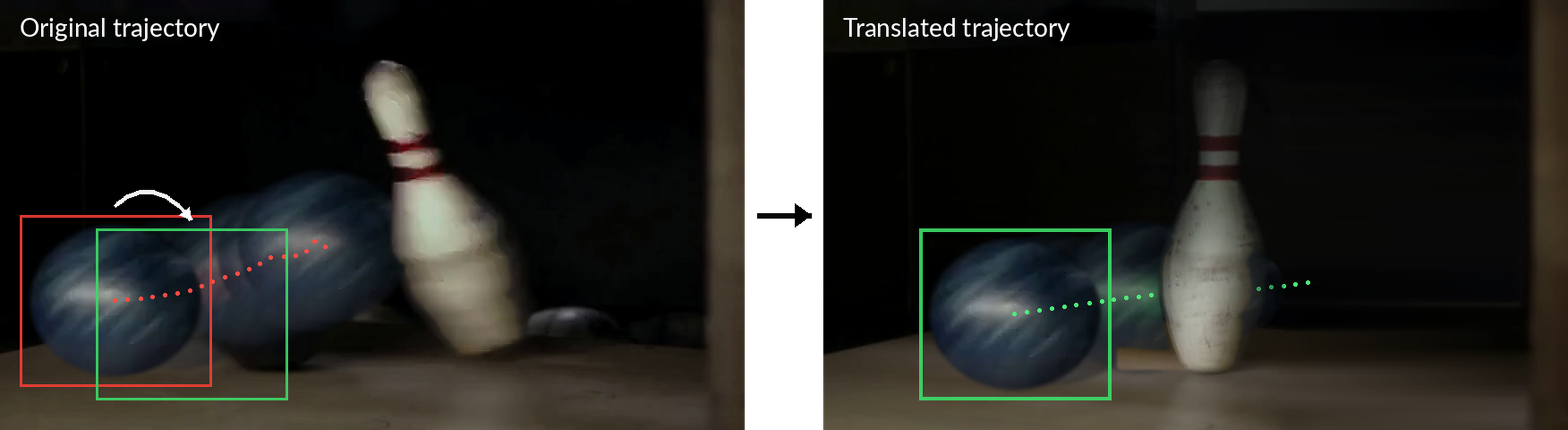}

    \captionof{figure}{%
    \textbf{\nameMethod} proposes a new video editing task:
    translating an object's 3D motion trajectory to a new starting
    location in the first frame while (i) otherwise preserving the
    video's identity and (ii) keeping both its motion and appearance
    plausible.
    The repeated translucent balls show the same ball at successive
    video frames, while the dotted lines trace its trajectory over time.
    Motion plausibility sometimes requires modifying the trajectory;
    for example, when translating the bowling ball's trajectory to miss
    the pin, its motion should no longer be deflected by the pin.
    We control the translation with two bounding boxes: the red box
    indicates the object's original starting location, the green box
    indicates its edited starting location, and the white arrow shows
    the requested translation.
    }
    \label{fig:teaser}

    \vspace{1.5em}
}{}{}
\makeatother

\maketitle

\begin{abstract}
Generative video editing has enabled creative control over an object's position in a video
by 
prescribing an object's 3D or 2D motion trajectory, while preserving both video plausibility and identity. 
However, manually specifying a plausible full motion trajectory, like the arcs of a bouncing ball, requires time and expertise, and may therefore not be a suitable editing task for non-experts or quick edits. In contrast, in the image domain, generative object translation has been established as a simple editing task that requires only a single drag to move an object to a new location; yet an equivalent method is still missing for videos.
We propose an analogous new editing task for videos that translates an object's 3D motion trajectory in a video while preserving identity and plausibility of appearance and motion, 
for example, translating the trajectory of a bouncing ball while preserving its relative motion.
The main challenge in training this task lies in obtaining paired video data for this scenario.
We introduce \nameDataset, a new data generation strategy for large-scale synthetic paired video data and a video generator \name fine-tuned with this data. We show that this enables generative movement of object trajectories.

\par\vspace{0.35em}
\noindent\centering
\textbf{Project Page: }
\href{https://chhatrekiran.github.io/trajectorymover/}
{chhatrekiran.github.io/trajectorymover}
\par
\vspace{-1em}

\end{abstract}

\section{Introduction}

Changing the motion of an object in an existing video is a fundamental video editing task. For example, an editor may want to change the trajectory of a basketball to enter a hoop or to bounce off the rim, without changing anything else in the video. Traditionally, changing an object's motion in a video requires re-shoots that may be expensive, time-consuming, or sometimes even impossible.

Recent advances in generative video editing resulted in several methods that allow changing the object motion in an existing video by specifying an exact 2D~\cite{wang2025ati} or 3D object trajectory~\cite{gu2025das}. However, while this detailed control is sometimes desirable, it has several drawbacks. First, manually specifying a full object trajectory in 2D or even 3D, such as a smooth parabola for a thrown object, is a time-consuming task that requires expertise. Second, it puts the responsibility for creating a trajectory that is plausible for the given scene on the shoulders of the user, rather than letting the prior of the generative model ensure its plausibility. This can be a difficult task, especially if the trajectory should capture interactions of the object with a scene: for example, what trajectory is realistic for a basketball bouncing off a rim?

Instead, we propose a much simpler fundamental task: can we translate the trajectory of an object in a video, while keeping its motion and scene interactions plausible? This affords a much simpler editing experience, where the user can simply drag a moving object in a video frame to a new position.
We introduce \name, a generative model that implements this task. Given a source video, the user annotates the first frame with two bounding boxes: a source bounding box around the object that is to be moved, and a target bounding box defining the target location. Our model then translates the trajectory to start at the target location, automatically adjusting the motion to preserve physically plausible scene interactions, but otherwise preserving the original trajectory relative to the new starting location.
For example, in Figure~\ref{fig:teaser} the trajectory of a bowling ball is moved to miss a pin, causing both the pin to remain standing, and the ball's trajectory to no longer be deflected by the pin.

The main challenge in training such a model is the lack of paired data for this task. While it is possible to annotate real-world videos by tracking objects, paired videos that show the same scene and the same object with translated trajectory are typically not available and would require an expensive controlled setup to create.
To address this challenge, we introduce a synthetic data generation strategy that automatically places new objects into existing scenes at locations that ensure object and trajectory visibility, procedurally synthesizes a variety of plausible trajectories for these objects to obtain sufficient motion diversity for good generalization, and uses a physics simulation to ensure plausible interaction between the objects and the scene. Note that we do not aim to teach the video generator about physics or plausible object motions, as this knowledge is already available in the video prior. The synthetic trajectories just need to be diverse enough to make the task that the video generator should perform clear. Additionally, we introduce strategy to reduce the number of interactions between the object and the scene in a subset of data samples using an online scene modification approach that removes parts of the scene that would block the object trajectory. This ensures that there is a subset of samples that demonstrates exact trajectory preservation, without the trajectory being modified by scene interactions.
Using this paired data, we formulate trajectory movement as a video-to-video generation task, encoding the bounding box control as an additional input frame. To preserve its generative prior, we fine-tune a video model~\cite{wan2025} by alternating between standard generation on real videos and our specific task. This strategy teaches the model to interpret the new control signal while relying on its pre-trained prior for realism and physical plausibility.

Extensive experiments demonstrate that \name translates trajectories more accurately and plausibly than methods requiring explicit trajectory specifications.

In summary, we propose the following contributions:
\begin{itemize}
    \item We identify a lack of simple-to-use editing tools for object trajectories in videos and propose trajectory movement as simple and fundamental tool to edit trajectories. We show that this approach addresses video editing tasks that are challenging with current tools.
    \item To address the lack of paired data, we introduce a synthetic data generation strategy to generate video pairs that share the same content, except for one object with moved trajectory and ensures trajectory visibility and motion diversity.
    \item We propose to formulate trajectory movement as a video-to-video generation task and carefully fine-tune a video generator without losing its original prior.
\end{itemize}

\section{Related Work}

\subsection{Conditional Video Generation and Editing}
Conditional video diffusion models extend base text or image conditioned architectures by incorporating auxiliary control signals.
Inspired by the success of ControlNet \cite{zhang2023adding} on controllable image generation, several approaches have introduced ControlNet-style hypernetworks to video synthesis \cite{gu2025diffusion,chen2023control,jeong2023ground,shi2024motion,burgert2025motionv2v,jiang2025vace,ouyang2024i2vedit}.
These methods adapt temporal mechanisms to enable guidance via diverse visual signals, such as depth maps, edge maps, and camera parameters.
Alternatively, other frameworks utilize source video input to facilitate video-to-video editing.
These approaches generally address tasks such as targeted appearance editing \cite{jiang2025vace,ju2025editverse,LucyEdit,ouyang2024i2vedit}, global stylization~\cite{ye2025stylemaster,li2026dreamstyle}, and novel view synthesis~\cite{bai2025recammaster,huang2025spacetimepilot,park2025redirector,jeong2025reangle,yu2025trajectorycrafter}.
To incorporate source guidance, many existing methods concatenate clean source video tokens with noisy target video tokens, either along the channel~\cite{LucyEdit,yu2025trajectorycrafter} or token dimension~\cite{bai2025recammaster,ju2025editverse,yu2025trajectorycrafter,lee2025generative}. Following this paradigm, \name integrates source video inputs via token concatenation.

\subsection{Motion-controlled Video Generation and Editing}
Recent generative video models have increasingly prioritized motion control, with sparse point tracks emerging as a flexible conditioning signal that can encode object motion, dense scene flow, or camera movement under a single representation~\cite{geng2025motion}. Many image-to-video (I2V) models allow trajectory conditioning but rely on a single reference frame~\cite{deng2024dragvideo,shi2024motion,wang2025ati}, neglecting the temporal context of the input video and losing the original scene information when motion is modified.
Existing video-to-video (V2V) approaches generally fall into two paradigms. Trajectory-based methods~\cite{burgert2025motionv2v,geng2025motion,lee2025generative,shin2025motionstream,deng2024dragvideo,teng2023drag, zhang2026flexcomposer, phung2026trace, hu2026preserverevealexpandfaithful, chang2026geometry, liu2025moright, zhou2026proxyup} condition generation on sparse point tracks for precise manipulation of object paths.
Concurrent work such as MotionV2V~\cite{burgert2025motionv2v} targets V2V editing via ``motion edits'' (deviations between source and target keypoint trajectories), CausualMotion~\cite{zhuang2026causalmotionstructuredphysicalreasoning} uses a combination of motion prompts and keyframes as input, and MotionStream~\cite{shin2025motionstream} achieves real-time streaming generation via sliding-window attention with attention sinks. Optical flow-based methods~\cite{cong2023flatten,koroglu2025onlyflow} instead leverage dense correspondence priors for detailed motion transfer.
While these works have demonstrated impressive control over object trajectories, they require exact specification of the trajectory. If moving an object trajectory results in new interactions between the object and the scene, a user needs to design any new motions resulting from these interactions manually, rather than letting the video prior generate them.

In this work, we address the complex task of shifting an object's 3D motion trajectory while preserving plausible interactions with the scene, without needing to manually specify detailed trajectories for updated interactions.
The primary obstacle to achieving such motion-preserving trajectory movement is the scarcity of paired video data suitable for supervised training.
We address this by introducing \nameDataset, a data generation strategy for large-scale synthetic data of paired videos specifically designed to provide the ground-truth supervision necessary for trajectory manipulation.
By fine-tuning a video generator on this diverse corpus, we achieve precise, generative control over object paths.

\begin{figure*}[t]
    \centering
    \includegraphics[width=\textwidth]{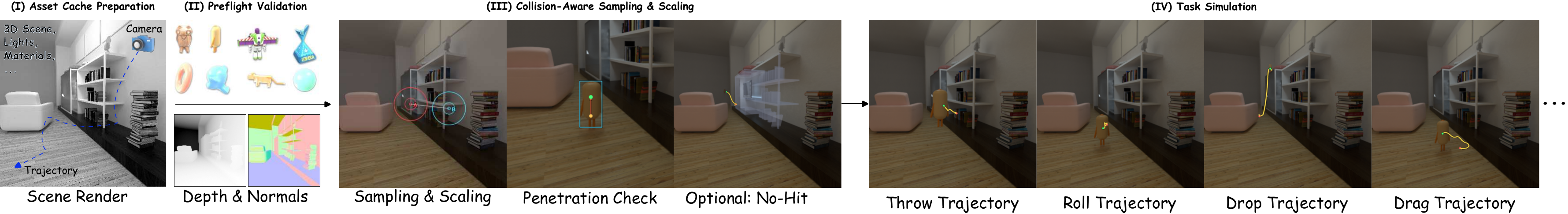}
    \caption{
    \textbf{\nameDataset data generation strategy.}
    Given a scene and a camera pose, we render the scene, use depth and normals of the render to place an object from an object library either in the air or on the ground, and generate a trajectory using a procedural path or an initial velocity that guides a physics simulation. In a video pair $(A, B)$, this strategy is repeated for both A and B, where the two have different first-frame placements, but share the same object and trajectory relative to the first-frame placement, except for any differences caused by different object-scene interactions.
    }
    
    \label{fig:data_pipe}
\end{figure*}

\section{Method}
Our goal is to generatively move the trajectory of an object in a video.
Conceptually, this requires removing the object from its original trajectory in the video and inserting it with a new trajectory, while 1) preserving the object's identity, 2) ensuring that the object follows the new trajectory, and 3) making sure the resulting video remains plausible, including interactions between the object and the containing scene. Note that 2) and 3) may be at odds -- for example, the moved trajectory of a thrown ball may pass through a wall. In this case, the object motion should remain plausible, for example, we would expect the ball to bounce off the wall. We want to provide simple controls for the user, thus we chose two bounding boxes as control signal (see Figure~\ref{fig:teaser}). The first bounding box selects an object in the first frame and the second describes its target position for the object, with the size of the bounding box providing control over the distance from the camera.

To achieve this, we propose a new synthetic data strategy \emph{\nameDataset} capable of generating a large video dataset that is aligned to this task, consisting of videos pairs that would be hard to find in any current real-world or synthetic datasets: video pairs that show the same video, except for one object with moved trajectory (Sec.~\ref{sec:data_gen}). 
Data generated by \nameDataset is used to fine-tune an existing diffusion-based video generator~\cite{wan2025} \emph{\name} that has a strong prior distribution over plausible videos (Sec.~\ref{sec:video_gen}).
To make sure the generator does not forget its prior, we use interleaved training that switches between unconditional generation of real videos and our task with synthetic videos. As conditioning strategy, we 
follow~\cite{bai2025recammaster} in treating the source video, the conditioning signal, and the target video as a single sequence, but only generate the tokens corresponding to the target video and keep the other tokens fixed.

\subsection{Data Generation Strategy}
\label{sec:data_gen}

Assume we are given a source video $A$ with $N_F$ frames where a foreground object follows a 3D trajectory $X = (x_i)_{i=1}^{N_F}$ with per-frame
object
positions $x_i \in \mathbb{R}^3$. We want to create a video $B$ where the same object instead follows a 3D trajectory $Y = (y_i)_{i=1}^{N_F}$ that is offset from $X$ by $\delta$: $y_i = x_i + \delta$, unless this would result in an implausible motion, like passing through walls. We
create pairs $(A,B)$ that can be used to fine-tune a video generator.
Additionally, we create binary segmentation videos $(M_A, M_B)$ for each video corresponding to the moved object; these can be used to create the 2D bounding boxes of the object in videos $A$ and $B$ that we need for our control signal. In this section, we describe our approach to create this dataset; an overview is shown in Fig.~\ref{fig:data_pipe}.

\paragraph{Scenes and objects.}
Our data generation approach takes as input a set of synthetic 3D scenes $\{S_i\}_{i=1}^{N_S}$ with a set of camera poses $\{C_i\}_{i=1}^{N_S}$, $C_i = c_1, c_2, \dots$ for each scene. Additionally, we assume a set of 3D objects $\{O_i\}_{i=1}^{N_O}$ to be inserted into the scenes is provided. To create a video pair $(A,B)$, we start by randomly choosing a scene $S_i$, a camera pose $c_j$ in the scene, and an object $O_k$.

\paragraph{Object trajectories.}
In addition to scenes and objects, we define a set of object trajectory types that the object should attempt to follow in videos $A$ and $B$, up to any scene interactions that may cause deviations from the trajectories. Assuming an initial placement $x_0$ of the object in the first frame:
\begin{itemize}
    \item \texttt{Static}: $x_0$ places the object on a supporting surface with zero initial velocity, resulting in zero motion.
    \item \texttt{Roll}: $x_0$ places the object on a supporting surface with a non-zero initial velocity, resulting in a rolling motion.
    \item \texttt{Drag}: $x_0$ places the object on a supporting surface with an elastic force that drags the object along a procedurally defined path (see supplement for details).
    \item \texttt{Drop}: $x_0$ places the object in the air with zero initial velocity, resulting in the object dropping down.
    \item \texttt{Throw}: $x_0$ places the object in the air with non-zero initial velocity, resulting in an arced trajectory.
    \item \texttt{Fly}: $x_0$ places the object in the air without gravity and with an elastic force that drags the object along a procedurally defined path (see supplement for details).
\end{itemize}
All procedural paths are defined relative to the initial position $x_0$. Both the parameters for the procedural paths and initial velocities are sampled randomly, resulting in a large variety of motions. See the supplement Section~\ref{sec:trajgendetails} for details on procedural paths and parameter distributions.
We simulate trajectories using robust rigid body physics~\cite{coumans2019pybullet} to ensure plausible interactions with scene geometry. As noted earlier, we do not aim to teach the video generator about physics or plausible object motions with these trajectories, as this knowledge is already available in the video prior. The trajectories just need to be diverse enough to make the task that the video generator should perform clear.

\paragraph{First-frame object placement.}
The initial placements $x_0$ and $y_0$ of the object in the source- and target videos need to be chosen such that (i) the object is clearly visible in the first frame, (ii) does not implausibly intersect any scene geometry, and (iii) allows for the chosen object trajectory type. We distinguish between two types of  placements: in the air and on the ground, depending on the chosen object trajectory type.
For air placements, we randomly choose a pixel of the first frame, shoot a ray from the camera through the pixel, and place the object at a random point along the visible part of the ray (before it hits the scene).
For ground placements, we randomly choose a pixel that correspond to a supporting surface (a pixel with an upward-pointing normal), shoot a ray through the pixel, and place the object so it rests on the point hit by the ray.
In both cases, we reject and re-sample placements that result in intersections between the object and the scene.
Given the initial positions, we can define $\delta = y_0 - x_0$.

In ground placements, the depth component of $y_0$ is well-defined in our 2D bounding box control signal as the depth where the object rests on the ground when placed inside the bounding box. For air placements, the depth needs to be derived from the size of the 2D bounding box. As the size of the bounding box may also depend on other factors, like the object rotation (for anisotropic object shapes), there is some ambiguity in the depth of $y_0$ for air placements.
To reduce this ambiguity, we
opt to choose pixels independently, but to correlate the depth of air placements in source and target videos. Specifically, we sample the depth in the target video using a Gaussian centered at the placement in the source video, resulting in predictable same-depth trajectory offsets $\delta$.

We choose object scale based on the screen size in the first frame, specifically by setting the maximum bounding box dimension to a target percentage of the frame height. We sample this percentage from two random settings: a close-up setting ($\mathcal{U}(20\%, 50\%)$) selected 22\% of the time, and a regular setting ($\mathcal{U}(7\%, 20\%)$) used for the remaining 78\%.

\paragraph{Online scene modification.}
Generating video pairs as described above results in plausible object trajectories. However, due to scene clutter, in many videos only a small part of the trajectory remains similar in a video pair. As soon as the object starts to interact with scene geometry, the trajectories in a video pair are likely to diverge. To get a more instructive set of videos that more clearly demonstrate our requirement of trajectory preservation, we modify the scene in a random subset of video pairs (roughly 50\%) so that larger parts of the trajectories in a video pair remain similar. Specifically, we remove any clutter in the scene that would intersect the projected trajectory in any of the two videos. Additional details are provided in the supplement.

\subsection{Video Generator}
\label{sec:video_gen}

Using data from \nameDataset (Section~\ref{sec:data_gen}), we train a video-to-video model that maps a source video $\sourcevideo \in \videospace$ to a target video $\targetvideo \in \videospace$, conditioned on an object displacement control signal $\controldisp \in \mathbb{R}^{\numchannel \times \height \times \width}$.
Here $\numf$, $\numchannel$, $\height$, and $\width$ denote the number of frames, color channels, and frame resolution.
The control signal $\controldisp$ encodes which object to move and where to place it in the target frame.

We use Wan2.1-T2V-1.3B~\cite{wan2025} as the DiT backbone and its VAE to map video frames $V$ to spatio-temporal latents $z \in \latentspace$.
We form three latent streams, $z_{\mathrm{trj}}$, $z_{\mathrm{src}}$, and $z_{\mathrm{bb}}$, and concatenate them along the temporal axis before denoising.
To match frame-level RoPE indexing~\cite{su2024roformer}, we place target frames first, from $0$ to $F'-1$, followed by source frames and the control latent.

\section{Implementation Details}

\begin{figure}[t]
    \centering
    \includegraphics[width=\linewidth]{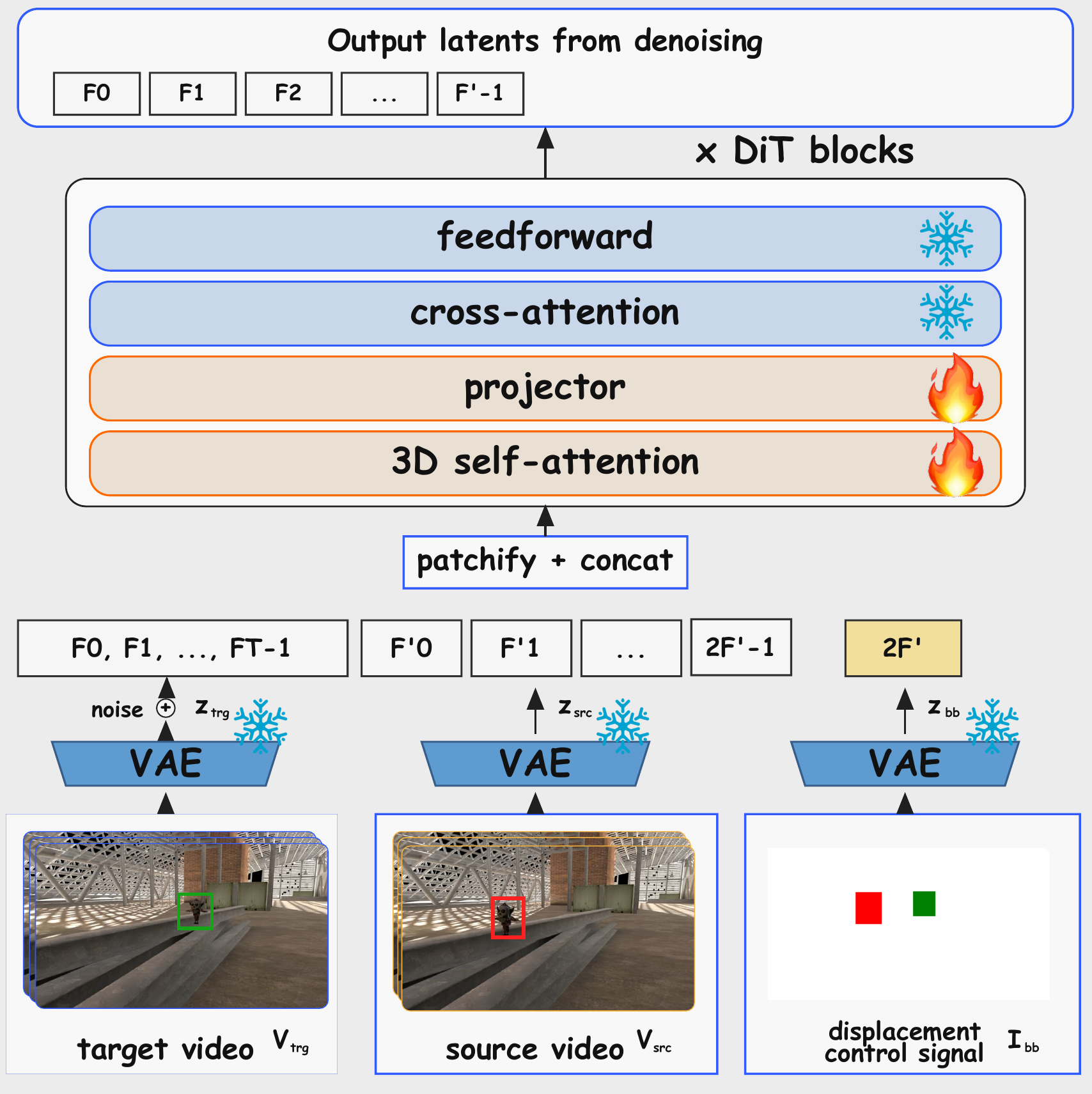}
    \caption{
        \textbf{\nameMethod architecture.}
        We concatenate three latent streams $z_{\mathrm{trj}}$, $z_{\mathrm{src}}$, and $z_{\mathrm{bb}}$ before denoising.
        In the control image, red marks the source box and green marks the target box.
    }
    \label{fig:arch}
\end{figure}

\paragraph{Data generation.}
\nameDataset utilizes Blender (Cycles) for rendering and PyBullet for physics. Our assets include curated Evermotion~\cite{evermotion_website} indoor scenes and 119 foreground objects (98 Objaverse models~\cite{objaverse} and 21 primitives). For each video pair, we sample scaled initial object placements, enforcing visibility, support, and intersection checks. We then simulate task-specific physics (drop, throw, roll, drag, etc.), optionally apply online scene modifications, and render the final videos.
For efficiency and reproducibility, we precompute collision meshes and save fixed seeds alongside full runtime configurations.

\paragraph{Training.}
Using bounding boxes extracted from ground-truth masks, we construct the input control signal $\controldisp$, where red and green boxes specify the object's initial location in the source and target sequences, respectively. To align with the native spatial and temporal resolution of the Wan2.1-T2V-1.3B backbone, frames are resized to 832 × 480, and sequences are limited to the first $\numf$ = 81 frames (yielding $F'$ = 21 latent frames). Crucially, we train the model using only empty (null) text conditions; as a result, users do not need to provide any text prompts at test time. As shown in Fig.~\ref{fig:arch}, we adopt a parameter-efficient tuning approach, optimizing only the self-attention and projector layers while freezing the rest of the network. To retain the original generative prior, we utilize a joint training scheme that alternates between the \nameMethod V2V task on our dataset and standard T2V generation on a large video corpus at a 7:3 ratio. The model is trained for 3,200 steps on an 8-H100-GPU machine with a total batch size of 16.

\section{Results}
\label{sec:results}

We evaluate \name{} on our trajectory movement task against several state-of-the-art video editing baselines, assessing identity preservation, trajectory adherence, and overall plausibility.

\paragraph{Dataset.}
We generate a total of $\sim33.3$k video pairs with \nameDataset that we split into $\sim31.6$k training pairs and $\sim1.6$k test pairs, where $106$ of our 3D scenes used for the training pairs and $7$ different scenes for the test pairs.
From the test pairs, we randomly choose 146 videos for evaluation.
All videos have a resolution of 1280$\times$720 and 81 frames at 16 fps. More detailed dataset information is provided in the supplement, Section~\ref{sec:dataset_statistics}.

\paragraph{Baselines.}
While no existing baseline supports our trajectory movement task out-of-the-box, we repurpose existing video generators to our task while keeping the core methods unchanged.
ATI~\cite{wang2025ati} moves objects to follow given 2D trajectories.
DaS~\cite{gu2025diffusion} moves objects by editing 3D tracking videos of the source object.
VACE~\cite{jiang2025vace} supports object movement through a bounding box based trajectory reference video.
I2VEdit~\cite{ouyang2024i2vedit} propagates the first frame edits to other frames, allowing us to move the object in the first frame.
SFM~\cite{liu2025shape} is a 3D aware video editing method that reconstructs an editable 3D mesh and uses it to guide motion editing. 
GWTF~\cite{burgert2025go} moves objects by warping the input noise with a given optical flow.

Since our data strategy is an important part of our contribution,
we do not retrain these baselines on our synthetic data; additionally, several of these methods are not directly compatible with our training data.
We do not include Motion Prompting~\cite{geng2025motion} or GenProp~\cite{liu2024generativevideopropagation}, as no code is available at the time of writing.
Most baselines require a full trajectory as input. To obtain a translated trajectory, we
extract a 3D trajectory from the source video by tracking the foreground object's estimated depth~\cite{video_depth_anything}. We then translate this trajectory to align with the target bounding box and convert it to each baseline's native format (see supplement for details). Notably, unlike our method, this naive translation fails to account for scene interactions.

\paragraph{Metrics.}
We evaluate each method on foreground identity preservation, background preservation, and trajectory adherence.
For each generated video, we extract foreground masks \(M\) with SAM3~\cite{carion2025sam3segmentconcepts}, while ground-truth masks \(M_{\text{gt}}\) come from dataset segmentations.
\(\text{SSIM}_{\text{bg}}\) measures background preservation by comparing generated frames to GT target frames only on background pixels.
We define foreground as the union of (i) \(M_{\text{gt}}\) and (ii) a slightly dilated \(M\), then define background as its complement; masked SSIM~\cite{wang2004image} is computed per frame and averaged over time.
\(\text{DINO}_{\text{fg}}\) measures foreground identity preservation: we take the first valid source-object crop using \(M_{\text{gt}}\) as reference, encode it with \texttt{dinov2-base}~\cite{oquab2023dinov2,jose2024dinov2meetstextunified}, and compute cosine similarity against generated-object crops using \(M\) across all visible frames, reporting the mean similarity as the final score.

Since our focus is trajectory movement, we additionally report trajectory-adherence metrics in image space using per-frame foreground bounding boxes from \(M\) and \(M_{\text{gt}}\).
\(\text{IoU}\) is the per-frame bounding-box IoU averaged over frames.
\(E_c\) is the per-frame bounding-box center distance averaged over frames.
\(E_s\) is the per-frame relative area error (with respect to GT area) averaged over frames.
\(E_v\) is the per-frame velocity \(L_2\) error averaged over frames.
\(R_s\) is the trajectory-start success rate, i.e. in what fraction of samples the start of the edited trajectory is correct, using a threshold of \(0.5\) on first-frame \(\text{IoU}\).
\(R_e\) is the trajectory-end success rate, using a threshold of \(20\) pixels on last-frame \(E_c\).
\(R_c\) is the collision-matching success rate: collisions are detected in generated trajectories as large direction changes (\(\ge 35^\circ\)) between consecutive frames; if corresponding GT direction changes are present and post-collision directions match with cosine similarity \(\ge 0.7\), the collision is counted as matched.

  \begin{table}[t]
  \centering
  \caption{
  \textbf{Comparison to Baselines on Synthetic Videos.}
  We compare background preservation (\(\text{SSIM}_\text{bg}\)), foreground identity preservation (\(\text{DINO}_\text{fg}\)), and trajectory adherence metrics of our generated video edits against baselines on 146
  test videos. Higher is better for all metrics except \(E_c\), \(E_s\), and \(E_v\), where lower is better.
  }
  \label{tab:comparison}
  \footnotesize
  \renewcommand{\arraystretch}{1.08}
  \setlength{\tabcolsep}{3pt}
  \begin{tabularx}{\linewidth}{
  >{\raggedleft\arraybackslash}p{1.05cm}
  >{\hsize=1.30\hsize\centering\arraybackslash}X
  >{\hsize=1.30\hsize\centering\arraybackslash}X
  >{\hsize=0.88\hsize\centering\arraybackslash}X
  >{\hsize=0.88\hsize\centering\arraybackslash}X
  >{\hsize=0.88\hsize\centering\arraybackslash}X
  >{\hsize=0.88\hsize\centering\arraybackslash}X
  >{\hsize=0.88\hsize\centering\arraybackslash}X
  >{\hsize=0.88\hsize\centering\arraybackslash}X
  >{\hsize=0.88\hsize\centering\arraybackslash}X
  }
  \toprule
  & \multicolumn{1}{c}{Bg. IP} & \multicolumn{1}{c}{Fg. IP} & \multicolumn{7}{c}{Trajectory Adherence} \\
  \cmidrule(lr){2-2}\cmidrule(lr){3-3}\cmidrule(lr){4-10}
  & \scriptsize $\mathrm{SSIM}_{\mathrm{bg}}$
  & \scriptsize $\mathrm{DINO}_{\mathrm{fg}}$
  & \scriptsize $\mathrm{IoU}$
  & \scriptsize $E_c\downarrow$
  & \scriptsize $E_s\downarrow$
  & \scriptsize $E_v\downarrow$
  & \scriptsize $R_s$
  & \scriptsize $R_e$
  & \scriptsize $R_c$ \\
  \midrule
  ATI     & \underline{0.62} & 0.32              & 0.12             & 120.9            & 5.91             & \underline{0.22} & 0.74             & 0.10             & 0.27             \\
  DaS     & 0.39             & \underline{0.37} & 0.14             & 133.7            & 6.69             & 0.24             & 0.77             & 0.12             & 0.29             \\
  VACE    & 0.59             & 0.18             & 0.15             & 136.2            & 6.51             & \textbf{0.18}    & 0.84             & 0.14             & 0.18             \\
  I2VEdit & 0.14             & 0.13             & 0.15             & 129.8            & 7.38             & 0.23             & \underline{0.89} & 0.09             & 0.23             \\
  SFM     & 0.50             & 0.23             & \underline{0.21} & \underline{90.4} & \underline{1.73} & 0.24             & 0.83             & \underline{0.20} & 0.18             \\
  GWTF    & 0.41             & 0.29             & 0.20             & 130.7            & 7.21             & 0.24             & 0.85             & 0.12             & \underline{0.34} \\
  \midrule
  Ours    & \textbf{0.85}    & \textbf{0.40}    & \textbf{0.23}    & \textbf{64.8}    & \textbf{1.39}    & 0.23             & \textbf{0.92}    & \textbf{0.24}    & \textbf{0.45}    \\
  \bottomrule
  \end{tabularx}
  \end{table}

\subsection{Comparison to Baselines}
\begin{figure*}[t]
    \centering
    \includegraphics[width=\textwidth]{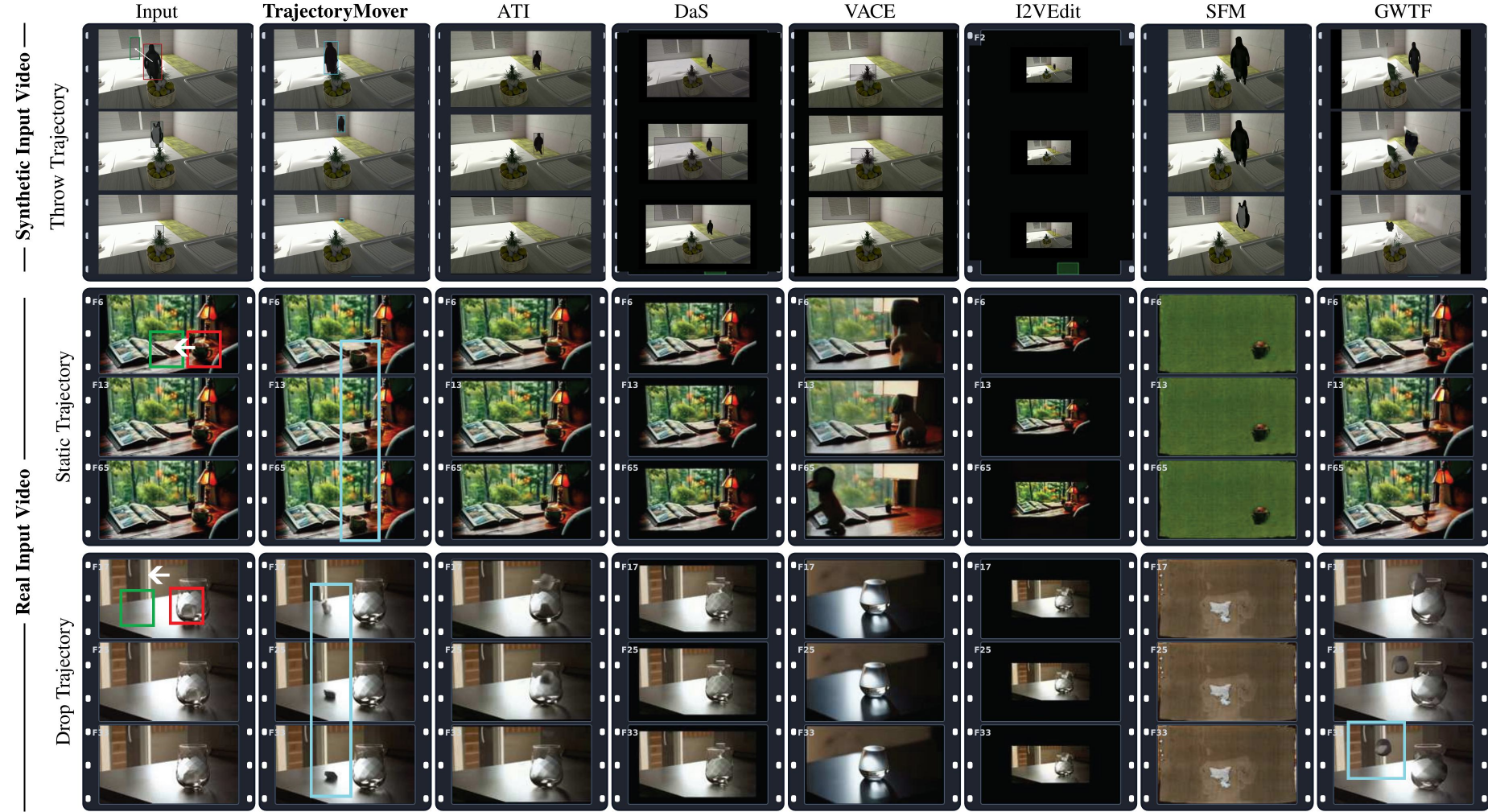}

  \caption{
      \textbf{Qualitative comparison with baselines.}
      We compare \name{} with SFM, ATI, DaS, VACE, GWTF, and I2VEdit on representative motion scenarios.
      Red boxes indicate the source object location in the input video, green boxes indicate the requested target location at
      frame 0, cyan boxes indicate the edited object region in generated results, and white arrows indicate the intended
      displacement direction.
      \name{} follows the intended motion most consistently while preserving object appearance and scene identity.
      Please zoom in for details.
  }

    \label{fig:baseline}
\end{figure*}

Table~\ref{tab:comparison} and Figure~\ref{fig:baseline} (top row) compare appearance preservation and trajectory adherence of \name to all baselines.
In Table~\ref{tab:comparison}, \name achieves the strongest overall performance, obtaining the best result on eight of the nine reported metrics.
Among the baselines, ATI provides the strongest background preservation ($0.62$), DaS the strongest foreground preservation ($0.37$), and SFM the strongest trajectory IoU ($0.21$) and center/scale accuracy.
VACE achieves the lowest velocity error ($0.18$), compared with $0.23$ for \name{}.
Overall, \name{} provides the strongest combined preservation of scene appearance, object identity, and requested trajectory behavior.

Figure~\ref{fig:baseline} (top row) shows a qualitative comparison on a test set sample. Additional qualitative results are available in the supplement (Section~\ref{sec:add_qual_comparison}).
\name{} tracks the intended trajectory more faithfully while maintaining foreground identity and background structure.
The translated object remains temporally coherent, exhibiting minimal texture collapse or shape drift, even as it adapts to novel scene depths and changing support surfaces.
Among the baselines, SFM is strongest at coarse trajectory following but suffers from artifacts introduced by its reconstruction and editing pipeline, resulting in poor object identity preservation.
ATI, DaS, and I2VEdit often do not follow the prescribed trajectory accurately,
while VACE and I2VEdit struggle with both trajectory following and object identity preservation.

All of these baselines aim to follow the moved trajectory even if the resulting motion conflicts with the scene geometry, thus making it implausible.
In contrast, a key advantage of \name{} is its ability to predict inherently plausible, scene aware motion directly from the source video and the simple trajectory relocation signal, without requiring a detailed manual trajectory at test time.
This is particularly evident in complex scenarios where an object must translate laterally while also naturally settling at a different depth or support surface.
Please zoom in on the figures for fine grained details. Additional qualitative examples, \nameDataset samples, and model result videos are provided in the supplementary material and supplementary video.

\subsection{User Study}
We conduct a blind pairwise perceptual study of motion plausibility on a subset of our synthetic test set.
For each test case, participants see two anonymized videos (A/B) from the same source-target setup and select the video with
more plausible object motion.
The A/B order is randomized per item, and method identities are recovered with a hidden answer key. For analysis, we report
preference rates and additionally fit a regularized Bradley--Terry model using iterative Luce Spectral
Ranking as implemented in \texttt{choix}~\cite{bradley1952rank,luce1959individual,maystre2015fast}.
The model is \(P(i \succ j)=\frac{e^{u_i}}{e^{u_i}+e^{u_j}}\), where \(u_i\) is the utility of method \(i\).
The participant study includes \(n=16\) participants with 25 pairwise judgments each (400 total votes).
As a complementary automatic evaluation, we also score plausibility with InternVL~\cite{chen2024internvl} over 10 runs (seeds 0--9) and report mean
\(\pm\) standard deviation. Results are summarized in Table~\ref{tab:user_real_ablation} (rows 1-7). On the synthetic
subset, \name{} ranks first in both participant and InternVL preference, with 82.75 participant preference and
0.63\(\pm\)0.13 InternVL preference. The strongest baselines remain substantially lower, with GWTF and SFM reaching 24.25
and 23.75 participant preference, respectively, and GWTF reaching 0.54\(\pm\)0.22 InternVL preference. Although the ordering
among baselines differs slightly between participants and InternVL, both consistently rank \name{} highest.

\begin{table}[t]
\centering
\caption{
\textbf{Synthetic user study, real-video evaluation, and ablations.}
Synthetic columns report participant and InternVL (IV) preference for the full model over each compared method or ablated
variant. Real columns report foreground identity preservation (DINO), target-start adherence ($R_s$), and IV
preference.
}
\label{tab:user_real_ablation}
\footnotesize
\renewcommand{\arraystretch}{1.05}
\setlength{\tabcolsep}{3.6pt}
\begin{tabularx}{\linewidth}{
>{\raggedright\arraybackslash}p{1.55cm}
>{\centering\arraybackslash}p{0.98cm}
>{\centering\arraybackslash}p{1.34cm}
>{\centering\arraybackslash}p{0.78cm}
>{\centering\arraybackslash}p{0.72cm}
>{\centering\arraybackslash}p{1.34cm}
}
\toprule
& \multicolumn{2}{c}{Synthetic} & \multicolumn{3}{c}{Real} \\
\cmidrule(lr){2-3} \cmidrule(lr){4-6}
Method & P Pref.$\uparrow$ & IV Pref.$\uparrow$ & DINO$\uparrow$ & $R_s\uparrow$ & IV Pref.$\uparrow$ \\
\midrule
ATI     & 14.75 & 0.40~$\pm$~0.21 & 0.32 & 0.47 & 0.32~$\pm$~0.11 \\
DaS     & 13.00 & 0.53~$\pm$~0.23 & 0.32 & 0.47 & 0.52~$\pm$~0.10 \\
VACE    & 15.25 & 0.52~$\pm$~0.07 & 0.24 & 0.47 & \underline{0.54~$\pm$~0.22} \\
I2VEdit & 17.25 & 0.38~$\pm$~0.11 & 0.14 & 0.46 & 0.49~$\pm$~0.28 \\
SFM     & \underline{23.75} & 0.50~$\pm$~0.10 & 0.24 & 0.47 & 0.45~$\pm$~0.64 \\
GWTF    & \underline{24.25} & \underline{0.54~$\pm$~0.22} & \underline{0.32} & 0.48 & \underline{0.54~$\pm$~0.22} \\
\midrule
Ours & \textbf{82.75} & \textbf{0.63~$\pm$~0.13} & \textbf{0.38} & \textbf{0.53} & \textbf{0.62~$\pm$~0.08} \\
\midrule
No mod.     & 60.75 & 0.56~$\pm$~0.10 & 0.37 & 0.49 & 0.58~$\pm$~0.10 \\
Coarse only & 66.75 & 0.59~$\pm$~0.12 & 0.35 & 0.48 & 0.56~$\pm$~0.11 \\
20\% data   & 71.75 & 0.61~$\pm$~0.14 & 0.33 & 0.44 & 0.53~$\pm$~0.12 \\
\bottomrule
\end{tabularx}
\end{table}

\subsection{Real Videos}
We compare our method to baselines on 37 non-synthetic web videos (sources: pixabay.com and pexels.com). Since no full
ground-truth trajectories are available for this set, we report foreground preservation (\(\text{DINO}
_\text{fg}\)), trajectory start adherence (\(R_s\)), and InternVL-based plausibility (Pref.) in
Table~\ref{tab:user_real_ablation}. Figure~\ref{fig:baseline} (last two rows) provides qualitative
comparisons on representative examples. While there is a gap to performance on synthetic data, \name{} shows promising
results, achieving the best foreground preservation while also better matching the intended target start position.
Specifically, \name{} obtains \(\text{DINO}_\text{fg}=0.38\) and \(R_s=0.53\), compared to the next best
values of 0.32 for foreground preservation and 0.48 for target-start adherence. InternVL also ranks \name{} first with
0.62\(\pm\)0.08 preference, followed by VACE and GWTF at 0.54\(\pm\)0.22.

\subsection{Ablation}
We ablate the effect of several key design choices in our data generation strategy on a model trained with the
resulting data. We train a model with data that has only primitives rather than a diverse set of objects from Objaverse
(only primitives). We ablate our choice to do online scene modifications in half of the video pairs by training on data
where all video pairs have been modified (only scene mod.) or none have been modified (w/o scene mod.). Finally, we ablate
the need for diverse trajectories using only the \texttt{Drop} trajectory (\texttt{Drop}-only). We
additionally ablate motion diversity by training on the smaller coarse motion set instead of the
expanded parameterized motion library used by \name{} (all), and we study training-data scale using 20\% and 50\% of the
full training set. Table~\ref{tab:ablation} and Table~\ref{tab:user_real_ablation} (last three rows) show that
all of the ablated choices contribute to trajectory adherence, plausibility, and identity preservation. A large set of objects is important for identity preservation (Primitives only drops $\text{DINO}_{\text{fg}}$ significantly) and
reducing motion diversity also lowers performance, showing that the expanded parameterized motion library
improves generalization to edited trajectories. Reducing training-data scale further degrades performance, with the
strongest drop at 20\% data and partial recovery at 50\%, though both remain below the full model. Background preservation
is not strongly affected by these design choices. Overall, the full model provides the best balance of trajectory adherence,
object identity preservation, and background consistency. A qualitative ablation is available in the supplement.

\begin{table}[t]
\centering
\caption{
\textbf{Ablation.}
We study foreground object diversity, online scene modification,
trajectory diversity, motion library, and training data scale.
M. denotes motion library. \emph{Coarse motions only} uses 9 coarse motions, while \name{} (all) uses
the expanded parameterized library 36 task variants.
}
\label{tab:ablation}
\footnotesize
\renewcommand{\arraystretch}{1.1}

\begin{tabularx}{\linewidth}{
@{}
c
@{\hspace{8pt}}
>{\raggedright\arraybackslash}X
@{\hspace{1pt}}
>{\centering\arraybackslash}p{1.18cm}
@{\hspace{6pt}}
>{\centering\arraybackslash}p{1.18cm}
@{\hspace{6pt}}
>{\centering\arraybackslash}p{1.18cm}
@{}
}
\toprule
&
&
\multicolumn{1}{c}{\scriptsize $\mathrm{SSIM}_{\mathrm{bg}}\uparrow$}
&
\multicolumn{1}{c}{\scriptsize $\mathrm{DINO}_{\mathrm{fg}}\uparrow$}
&
\multicolumn{1}{c}{\scriptsize $\mathrm{IoU}_{\mathrm{traj}}\uparrow$}
\\
\midrule

\multirow{4}{*}{\rotatebox{90}{\hspace{4pt}\textbf{Design}}}
& Primitives only
& \textbf{0.93} & 0.15 & 0.20 \\
& Scene mod. only
& \underline{0.92} & \underline{0.44} & 0.16 \\
& No scene mod.
& \underline{0.92} & \underline{0.44} & 0.17 \\
& \texttt{Drop} only
& \underline{0.92} & \textbf{0.45} & \underline{0.21} \\

\midrule

\multirow{1}{*}{\rotatebox{90}{\hspace{4pt}\textbf{M.}}}
& Coarse motions only
& 0.91 & 0.41 & 0.18 \\

\midrule

\multirow{2}{*}{\rotatebox{90}{\hspace{4pt}\textbf{Scale}}}
& 20\% data
& 0.90 & 0.39 & 0.14 \\
& 50\% data
& 0.91 & 0.42 & 0.19 \\

\midrule

& Ours
& \underline{0.92} & \textbf{0.45} & \textbf{0.27} \\

\bottomrule
\end{tabularx}
\end{table}

\subsection{Limitations}

Although \name{} substantially improves trajectory movement over existing
baselines, precise long-horizon control remains challenging.
On the 146 video synthetic evaluation in Table~\ref{tab:comparison},
\name{} achieves strong trajectory-start success
($R_s=0.92$), but lower trajectory-end success ($R_e=0.24$) and
collision-matching success ($R_c=0.45$), with
$\mathrm{IoU}=0.23$ and an average center error of $64.8$ pixels.
This indicates that the model reliably follows the requested relocation
at the beginning of the video, but can gradually deviate from the target
motion over time, particularly when interactions or occlusions make the
future trajectory ambiguous.

Generalization to unconstrained real videos also remains weaker than on
synthetic data.
On the real videos in Table~\ref{tab:user_real_ablation},
\name{} retains the strongest foreground preservation and plausibility,
but target start adherence decreases to $R_s=0.53$, compared with
$0.92$ on the synthetic evaluation.
Improving robustness to real world appearance variation, depth
ambiguity, occlusion, and more complex multi object dynamics remains
important future work.

\section{Conclusion}

In this work, we presented \name, the first framework to move an object's 3D motion trajectory in a video.  
To overcome the inherent lack of paired training data, we introduced \nameDataset, a synthetic data generation strategy that leverages physics simulations and online scene modification to create plausible, diverse trajectory-shifted video pairs.
By formulating this task as a video-to-video generation problem and employing a balanced fine-tuning strategy, our model successfully preserves the generative prior of the original backbone while learning complex spatial transformations.

An interesting avenue for future research could be adding additional control options for object trajectories. For example, can we use an arbitrary frame instead of only the first frame to move a trajectory? Can we rotate, scale, or speed up trajectories? Can we transfer trajectories between different objects? Can we simulate changing the mass of objects? Our work can be seen as an early example of the more general topic of motion preservation, analogous to identity preservation. Can we preserve the motion of non-rigid articulated objects, like characters, while changing the motion of only one limb? Can we re-target an existing character motion to a new environment or scene? Several of these question remain largely underexplored and we hope that our work inspires new research in this direction.

\section{Acknowledgments}
We thank Valentin Deschaintre and Iliyan Georgiev for insightful discussions and valuable feedback. We are also grateful to Yannick Hold-Geoffroy and Vladimir Kim for their help with the assets used in dataset generation, and to Zhening Huang for support with the model pipeline. The core ideas for this project were developed while Kiran Chhatre was an intern at Adobe Research.

\clearpage

{
    \small
    \bibliographystyle{ieeenat_fullname}
    \bibliography{main}
}
\appendix
\clearpage

\setcounter{section}{0}
\setcounter{subsection}{0}

\counterwithin{figure}{section}
\counterwithin{table}{section}

\renewcommand{\thefigure}{\thesection.\arabic{figure}}
\renewcommand{\thetable}{\thesection.\arabic{table}}

\clearpage
\maketitlesupplementary

\section{Overview}
\label{sec:suppl_overview}

This supplementary material includes the following parts:
\begin{enumerate}[leftmargin=*, align=left, labelwidth=2.5em,
                  labelsep=0.5em, itemindent=0pt, labelindent=10pt]

    \item[\textbf{Sec.~\ref{sec:interaction_eval}:}]
    3D trajectory and interaction evaluation.
    We evaluate the reconstructed 3D trajectory, rather than evaluating the 2D trajectory in image space. Additionally, we evaluate the accuracy of object-scene interactions,
    using trajectory and contact-based metrics.

    \item[\textbf{Sec.~\ref{sec:backbone_generalization}:}]
    Generalization across video backbones.
    We fine tune an LTX-Video 0.9.7 backbone using the same
    trajectory movement supervision and compare it with the Wan based
    model.

    \item[\textbf{Sec.~\ref{sec:supp_baseline_repurpose}:}]
    Baseline repurposing details.
    We describe how we adapt the baselines ATI, DaS, VACE, I2VEdit,
    SFM, and GWTF in our comparison and discuss excluded baselines as well as limitations of simple trajectory transfer.

    \item[\textbf{Sec.~\ref{sec:trajgendetails}:}]
    \nameDataset details.
    We provide all details of the data generation procedure, including
    object trajectory simulation, initial velocity and procedural-path
    sampling, online scene modification, implementation details, and
    dataset statistics.

    \item[\textbf{Sec.~\ref{sec:add_qual_comparison}:}]
    Additional qualitative comparisons.
    We provide additional comparisons with all evaluated baselines on
    synthetic and real videos.

    \item[\textbf{Sec.~\ref{sec:qual_ablation}:}]
    Qualitative ablation examples.
    We visualize representative results for several ablations reported
    in Table~3 of the main paper.

    \item[\textbf{Sec.~\ref{sec:user_study_interface}:}]
    User study interface.
    We show the interface and evaluation criteria used for the
    participant study.
\end{enumerate}

\begingroup
\renewcommand{\thefootnote}{\ddag}

In addition, our project webpage\footnotemark{} provides qualitative videos
showing ground-truth source--target pairs, comparisons between \name{} and
all baseline methods, and comparisons between \name{} and its ablation variants.

\footnotetext{\href{https://chhatrekiran.github.io/trajectorymover/}
{chhatrekiran.github.io/trajectorymover}}
\endgroup

\section{3D Trajectory and Interaction Evaluation}
\label{sec:interaction_eval}

Table~1 of the main paper evaluates trajectory adherence in image
space. Here, we instead evaluate whether an edited trajectory
(i) preserves the source relative 3D motion when no new interaction
requires changing it, and
(ii) adapts appropriately when relocation changes the interaction
between the object and the scene. We evaluate the same 146 synthetic test videos as
Table~1 of the main paper, partitioned using the simulator logs into
74 no interaction (No Int.) and 72 interaction (Int.) examples.
For the synthetic source and target videos, per frame 3D object
trajectories and interaction events are obtained from Bullet
and serve as ground truth.
For generated RGB videos, simulator state is unavailable, so we
reconstruct an approximate 3D trajectory $\hat{\mathbf p}_t$ by
combining SAM3 foreground masks, estimated video depth, and the known
camera parameters. Because the estimated video depth is not metrically scaled, we align
its scale to the rendered scene depth using visible background regions
before back projecting the object trajectory with the known camera
parameters.
Thus, the reported 3D errors also reflect uncertainty introduced by
the reconstruction. All generated methods are evaluated using the same
pipeline.

We compare two VACE control settings.
VACE-\emph{Rigid} receives the source trajectory translated to the
requested first frame target position. It therefore explicitly follows the translated source trajectory but
cannot account for trajectory changes caused by newly encountered
scene geometry.
VACE-\emph{Oracle}$^\ast$ instead uses its native trajectory control
constructed from the exact GT target trajectory exported from Bullet and is included only as an upper bound.
\name{} receives no future trajectory and is conditioned
only on the source video and the source and target boxes in the first
frame.

Let $\hat{\mathbf p}_t$, $\mathbf p_t^{gt}$, and
$\mathbf p_t^{src}$ denote the reconstructed generated, GT target,
and source 3D object positions, respectively, and let $D$ denote the
3D object size used for normalization.
Normalized Average Displacement Error (N-ADE) measures the
mean 3D distance to the GT target trajectory,
$\frac{1}{TD}\sum_{t=1}^{T}
\|\hat{\mathbf p}_t-\mathbf p_t^{gt}\|_2$, while
Normalized Final Displacement Error (N-FDE) measures the
corresponding error at the final frame,
$\|\hat{\mathbf p}_T-\mathbf p_T^{gt}\|_2/D$.
To evaluate preservation of the source motion independently of its
new starting position, Relative Motion Error
(Rel.-Motion Err.) measures
$\frac{1}{|\mathcal T|D}\sum_{t\in\mathcal T}
\|(\hat{\mathbf p}_t-\hat{\mathbf p}_0)-
(\mathbf p_t^{src}-\mathbf p_0^{src})\|_2$.
For interaction examples, $\mathcal T$ contains the frames before the
first GT target interaction. For no interaction examples, it contains
the full sequence.
For interaction examples, we additionally report
Contact Success, the fraction of examples in which the expected
GT object scene contact is reproduced. A generated contact is detected when the estimated distance between the
object surface and the corresponding scene geometry is below
$\tau=0.05D$.
We also report Post Contact Direction Error (Post-Dir.), the
cosine distance between the generated and GT 3D motion directions after
contact, with each post contact direction computed from the displacement
over the following $k=3$ frames.

Table~\ref{tab:interaction3d} evaluates trajectory adaptation both when
the translated source motion remains valid and when relocation changes
the object scene interaction.
For the no interaction subset, VACE-\emph{Rigid}, which is explicitly
conditioned on the translated trajectory, is slightly more accurate,
with N-ADE/N-FDE of 0.56/0.79 compared with 0.60/0.82 for \name{},
and Rel.-Motion Err.\ of 0.18 versus 0.20.
Thus, when no adaptation is required, \name{} closely preserves the
source relative motion despite receiving no future trajectory.

When relocation changes the interaction, rigidly translating the
source trajectory is no longer sufficient.
Compared with VACE-\emph{Rigid}, \name{} reduces Rel.-Motion Err.\
from 0.30 to 0.21, N-ADE from 1.10 to 0.76, and N-FDE from
1.54 to 1.01, while increasing Contact Success from 0.58 to 0.74
and reducing Post-Dir.\ from 0.55 to 0.34.
VACE-\emph{Oracle}$^\ast$ achieves the best values, as expected,
because it receives the GT target trajectory.




\begin{table}[t]
\centering
\caption{
\textbf{3D trajectory and interaction adaptation.}
VACE-\emph{Rigid} is conditioned on the source trajectory translated
to the requested target start, while VACE-\emph{Oracle}$^\ast$ receives
the GT target trajectory from the simulator.
Oracle$^\ast$ uses GT future motion and is included only as an upper bound.
\name{} receives no future trajectory.
Bold indicates the better method between VACE-\emph{Rigid}
and \name{}.
}
\label{tab:interaction3d}
\footnotesize
\renewcommand{\arraystretch}{1.08}
\setlength{\tabcolsep}{3.5pt}

\begin{tabular}{clccc}
\toprule
&
&
\textbf{VACE}
&
\textbf{VACE}
&
\textbf{\name{}}
\\[-2pt]
&
&
\emph{Rigid}
&
\emph{Oracle$^\ast$}
&
\emph{None}
\\
\midrule

\multirow{3}{*}{\rotatebox{90}{\textbf{No Int.}}}
& Rel.-Motion Err. $\downarrow$
& \textbf{0.18} & 0.11 & 0.20 \\
& N-ADE $\downarrow$
& \textbf{0.56} & 0.33 & 0.60 \\
& N-FDE $\downarrow$
& \textbf{0.79} & 0.47 & 0.82 \\

\midrule

\multirow{5}{*}{\rotatebox{90}{\textbf{Int.}}}
& Rel.-Motion Err. $\downarrow$
& 0.30 & 0.17 & \textbf{0.21} \\
& N-ADE $\downarrow$
& 1.10 & 0.58 & \textbf{0.76} \\
& N-FDE $\downarrow$
& 1.54 & 0.82 & \textbf{1.01} \\
& Contact Succ. $\uparrow$
& 0.58 & 0.88 & \textbf{0.74} \\
& Post-Dir. $\downarrow$
& 0.55 & 0.24 & \textbf{0.34} \\

\bottomrule
\end{tabular}

\end{table}

\section{Generalization Across Video Backbones}
\label{sec:backbone_generalization}

To evaluate whether \name{} depends on the particular generative prior
used in the main experiments, we additionally fine-tune
LTX-Video 0.9.7~\cite{HaCohen2024LTXVideo} on the same
trajectory-movement task and \nameDataset training data.
We use the same source-video and first-frame source/target box
conditioning as for the Wan2.1 model, adapting only the
architecture-specific input and fine-tuning implementation required by
the LTX-Video 0.9.7 backbone.

Table~\ref{tab:backbone_generalization} compares the two backbones on
the same evaluation set using representative appearance-preservation
and trajectory-adherence metrics.

The LTX-Video 0.9.7 model retains similar foreground identity and trajectory
adherence, while showing a larger drop in background preservation and
target-start accuracy.
This indicates that the trajectory-movement formulation and
\nameDataset supervision transfer across different video-generation
backbones, although the final quality remains influenced by the
underlying generative prior.

\begin{table}[t]
\centering
\caption{
\textbf{Generalization across video backbones.}
We fine-tune two pretrained video generators using the same
trajectory-movement supervision.
Both backbones learn the proposed control, while the final generation
quality depends on the underlying video prior.
}
\label{tab:backbone_generalization}
\footnotesize
\renewcommand{\arraystretch}{1.08}
\setlength{\tabcolsep}{4pt}

\begin{tabularx}{\linewidth}{
>{\raggedright\arraybackslash}X
>{\centering\arraybackslash}X
>{\centering\arraybackslash}X
>{\centering\arraybackslash}X
>{\centering\arraybackslash}X
}
\toprule
Backbone
& $\text{SSIM}_{\text{bg}}\uparrow$
& $\text{DINO}_{\text{fg}}\uparrow$
& $\text{IoU}_{\text{traj}}\uparrow$
& $R_s\uparrow$
\\
\midrule
Wan-T2V
& \textbf{0.85}
& \textbf{0.40}
& \textbf{0.23}
& \textbf{0.92}
\\
LTX Video
& 0.79
& 0.37
& 0.21
& 0.86
\\
\bottomrule
\end{tabularx}
\end{table}

\section{Baseline Repurposing Details}
\label{sec:supp_baseline_repurpose}

\paragraph{Setup}
No baseline natively supports our source to target object relocation task, so we repurpose each baseline through external controls while keeping the original model architecture and training procedure unchanged.
Each video pair we apply the baselines to provides source video $A$, ground truth target video $B$, and source/target segmentation videos where the foreground object is black and the background is white.
For prompt conditioned methods, we use the same neutral prompt across methods.
We normalize all baseline outputs to a common evaluation format using letterbox resizing, preserving aspect ratio to $1280\times720$, followed by temporal resampling to 81 frames at 16 fps. Temporal normalization is performed by uniformly sampling 81 target positions on the source video timeline and assigning each target position its nearest source frame, which reuses frames when the original video is shorter.

\begin{figure*}[t]
    \centering
    \includegraphics[width=.97\linewidth]{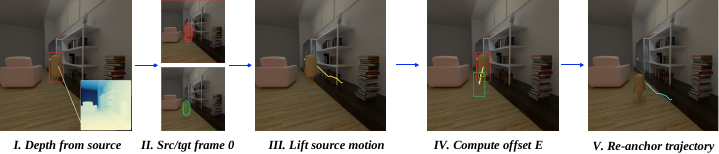}
    \caption{
        \textbf{Common baseline repurposing pipeline.}
        We convert each source-target case into method specific controls.
        We estimate source depth, extract source and target frame 0 masks, lift source object motion to a 3D trajectory proxy, compute the frame 0 displacement \(E\), and re-anchor the trajectory to the target start.
        Red indicates source localization, green indicates target localization, and trajectory overlays visualize source and re-anchored motion elements used for downstream baseline control conversion.
    }
    \label{fig:repurpose_pipeline}
\end{figure*}

\paragraph{Common 3D trajectory extraction for motion conditioned baselines}
ATI, DaS, and VACE require some form of trajectory-like control as input condition, so for these baselines we reconstruct a 3D object trajectory from the source video.
As shown in Fig.~\ref{fig:repurpose_pipeline}, we first estimate source depth from $A$ using Video-Depth-Anything~\cite{video_depth_anything}. As this per-frame depth includes correspondences across the video, we can simply select one or multiple sample points in the first-frame mask of the source object, get their depth, and follow them across the video to obtain 3D object trajectories. We then move these 3D trajectories with the ground truth edit offset $\delta$, giving us 3D object trajectories for the edited video.
These moved 3D trajectories are converted into each baseline’s native control format.
Note that these moved trajectories cannot take into account interactions with the scene, which is a natural limitation of existing methods that require exact trajectories as input condition.

\paragraph{ATI}
ATI~\cite{wang2025ati} is driven through its native trajectory interface. We project the moved 3D trajectories back to the 2D image plane of the target camera to obtain 2D trajectories that we can input directly into ATI.
We optionally add a small number of static background anchors to make sure the camera remains static.
ATI inference is then run normally with the target frame 0 image and generated tracks. The original output resolution is $832\times464$, at 16 fps, with 81 frames over 5 s.

\paragraph{DaS}
We obtain 3D tracks for the source video using one of the tracking methods used by DaS~\cite{gu2025diffusion}, apply the edit offset $\delta$ to the tracks of the foreground object, and feed the edited tracks, as well as the source video as input conditions to DaS.
This is fully compatible with the existing DaS pipeline. The original output resolution is $720\times480$, at 9.68 fps, with 49 frames over 5 s.

\paragraph{VACE}
VACE~\cite{jiang2025vace} is repurposed using its trajectory video control interface.
We create bounding boxes of the edited foreground object using the per-frame depth point cloud of the source object moved with the edit offset $\delta$ and use these bounding boxes to render a native layout trajectory control video.
We additionally provide (i) a source object reference crop from source segmentation and (ii) background preserving controls marking editable regions.
These controls are passed to the original VACE inference. The original output resolution is $832\times480$, at 16 fps, with 81 frames over 5 s.

\paragraph{I2VEdit}
I2VEdit~\cite{ouyang2024i2vedit} is repurposed via first-frame editing.
We extract the source object from frame 0 using source segmentation, remove the original object region with LaMa~\cite{suvorov2021resolution} inpainting using a dilated removal mask, and paste the object at the target frame 0 location using geometric compositing.
The edited first frame is then used with the standard I2VEdit pipeline to propagate edits over time. The original output resolution is $512\times512$, at 8 fps, with 40 frames over 5 s. After normalization for fair comparison, the result appears visually smaller within the padded frame because we preserve aspect ratio rather than stretching the lower-resolution, shorter raw output.

\paragraph{SFM}
Shape-for-Motion (SFM)~\cite{liu2025shape} is a 3D-aware video editing framework executed through its original multi-stage reconstruction, edit, and render pipeline.
Unlike ATI, DaS, and VACE, SFM does not use our depth-based trajectory transfer control.
Instead, for relocation we apply an external geometric edit by translating the reconstructed canonical mesh using the source and target frame 0 mask displacement.
The main SFM model remains unchanged. The original output resolution is $768\times512$, at 7 fps, with 21 frames over 3.0 s.

\paragraph{GWTF}~\cite{burgert2025go} moves objects by warping the input noise with a given optical flow. Out-of-the-box, it only allows specifying object trajectories where the first frame is unchanged. We adapt this to our setting by constructing per-case translated trajectory templates from source motion tracks, anchored so the object starts at the target location in the first frame. We then condition I2V generation on edited first-frame inputs while using the translated warped-noise trajectory.

\paragraph{Other excluded baselines.}
We do not include GenProp~\cite{liu2024generativevideopropagation} because public code is not available.
We also do not include VideoHandles~\cite{Koo:2025VideoHandles}, since it is designed for static object composition editing in static scenes and is not directly suitable for moving object trajectory relocation.

\paragraph{Limitation of simple trajectory transfer.}
For ATI, DaS, and VACE, we use copied source motion re-anchored to the target start position.
By design, this transfer does not explicitly reason about new scene interactions encountered after relocation, which can reduce plausibility under collisions.

\section{\nameDataset Details}
\label{sec:trajgendetails}

Algorithm~\ref{alg:trajatlas_datagen} summarizes the high level
\nameDataset generation procedure. For each sample, we select a scene,
camera, foreground object, motion task, and valid source and target
first-frame placements. For no hit configurations, we estimate the
expected motion path and remove intersecting non structural scene
objects before simulation. We then simulate the source and target
motions using the same task parameters, render the corresponding RGB
videos and foreground masks, and retain pairs that satisfy our dataset
validity checks.

\begin{algorithm}[t]
\caption{\nameDataset data generation}
\label{alg:trajatlas_datagen}
\small
\begin{algorithmic}
\Require Scenes $\mathcal{S}$, objects $\mathcal{O}$, tasks $\mathcal{T}$
\Ensure Paired dataset $\mathcal{D}$
\State $\mathcal{D}\gets\emptyset$
\For{each sample}
    \State Sample scene, camera, object, task, and source/target starts
    \If{no-hit}
        \State Remove non structural obstacles along the expected path
    \EndIf
    \State Simulate and render source/target videos and masks
    \If{valid pair}
        \State Add pair to $\mathcal{D}$
    \EndIf
\EndFor
\State \Return $\mathcal{D}$
\end{algorithmic}
\end{algorithm}

\subsection{Object trajectory details}
\label{sec:object_trajectory_details}
\begin{figure}[t]
    \centering
    \includegraphics[width=\linewidth]{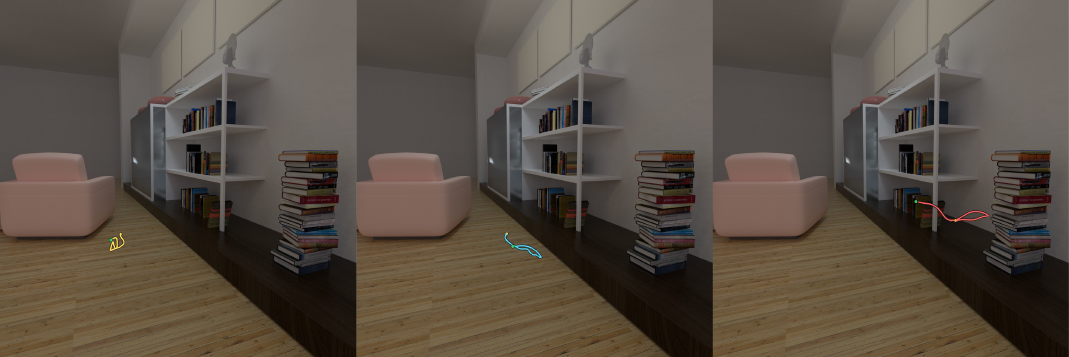}
   \caption{
\textbf{Example drag trajectory variants.}
We visualize three planar drag references used in our simulator:
spiral (left), circular (middle), and S-shaped (right), rendered from
the scene camera as 3D trajectories.
Each curve shows directed start-to-goal progression and illustrates
the procedural motion templates used for drag supervision.
}
    \label{fig:drag_variants}
\end{figure}

We simulate all object motions with rigid body dynamics in Bullet~\cite{coumans2019pybullet} and produce our trajectories via initial velocities, gravity, and/or an additional elastic force that we add to the simulation to drag the object along a procedurally defined path, as shown in~\ref{fig:drag_variants}. Each trajectory starts from a valid first-frame placement $x_0$ that is visible from the camera and intersection-free.

\paragraph{Initial velocity sampling.}
For motion types initialized by an impulse, we sample the initial velocity from motion-specific distributions.
For \texttt{Throw}, the horizontal speed is sampled from a three-component Gaussian mixture:
\begin{equation}
\begin{aligned}
v_{xy}^{\mathrm{throw}} \sim{}&
0.53\,\Normal{3.533}{0.624} \\
&+0.35\,\Normal{1.933}{0.330} \\
&+0.12\,\Normal{1.267}{0.206}.
\end{aligned}
\end{equation}
The vertical speed and azimuth are sampled as
\begin{equation}
v_z^{\mathrm{throw}} \sim \Normal{0.733}{0.429},
\end{equation}
\begin{equation}
\theta^{\mathrm{throw}} \sim \Uniform{-175^\circ}{175^\circ}.
\end{equation}

For \texttt{Roll}, the horizontal speed is sampled as
\begin{equation}
\begin{aligned}
v_{xy}^{\mathrm{roll}} \sim{}&
0.53\,\Normal{1.50}{0.334} \\
&+0.35\,\Normal{0.583}{0.156} \\
&+0.12\,\Normal{0.203}{0.056},
\end{aligned}
\end{equation}
with zero vertical velocity,
\begin{equation}
v_z^{\mathrm{roll}} = 0,
\end{equation}
and azimuth
\begin{equation}
\theta^{\mathrm{roll}} \sim \Uniform{-175^\circ}{175^\circ}.
\end{equation}

\paragraph{Procedural path sampling.}
For path-controlled motions, we procedurally generate a reference trajectory and move a target point along it.
The simulated object is pulled toward this moving target using a damped spring controller, with an additional orientation term that aligns the object to the target orientation.
For \texttt{Drag} motions, we project the sampled path onto the ground plane to remove vertical components before simulation.

We use S-shaped, O-shaped, U-shaped, C-shaped, 8-shaped, Z-shaped, spiral, sinusoidal, falling-leaf, pendulum, and door-swing paths.
The falling-leaf and pendulum paths are used only for airborne motions.
The door-swing path models rotation around a vertical hinge axis located on one side of the object.

The target motion along a path is controlled by its speed, start position, and end position.
The path speed is sampled as
\begin{equation}
\begin{aligned}
s \sim{}&
0.53\,\Normal{0.95}{0.309} \\
&+0.35\,\Normal{0.40}{0.114} \\
&+0.12\,\Normal{0.167}{0.051}.
\end{aligned}
\end{equation}
The start position along the path, expressed as a fraction of path length, is sampled as
\begin{equation}
\begin{aligned}
u_{\mathrm{start}} &\sim \Normal{\mu_s}{0.29}, \\
\mu_s &\sim \Uniform{0}{0.5},
\end{aligned}
\end{equation}
and the end position is sampled as
\begin{equation}
u_{\mathrm{end}} \sim \Normal{0.78}{0.057}.
\end{equation}

All procedural paths share orientation, extent, and curvature parameters.
The path orientation is sampled relative to the camera azimuth:
\begin{equation}
\phi \sim \Uniform{-175^\circ}{175^\circ}.
\end{equation}
The path extent, defined as the largest side of the path bounding box, is sampled as
\begin{equation}
\begin{aligned}
e &\sim \Normal{\mu_e}{0.17}, \\
\mu_e &\sim \Uniform{0.85}{1.45}.
\end{aligned}
\end{equation}
The curvature parameter controls the amplitude of curved paths, including the sinusoid amplitude, the width of the 8-shaped path, and the arc height of the U-shaped path:
\begin{equation}
\kappa \sim \Normal{0.31}{0.09}.
\end{equation}

Some path families have additional parameters.
For the pendulum path, we sample length and swing angle as
\begin{equation}
\begin{aligned}
\ell &\sim \Normal{0.85}{0.17}, \\
\alpha &\sim \Normal{48^\circ}{13.9}.
\end{aligned}
\end{equation}
For the door-swing path, we sample radius, rotation angle, and hinge height as
\begin{equation}
\begin{aligned}
r &\sim \Normal{0.575}{0.159}, \\
\beta &\sim \Normal{66.5^\circ}{18.2}, \\
h &\sim \Normal{0.27}{0.087}.
\end{aligned}
\end{equation}

\subsection{Online scene modification details}
The goal of this step is to reduce scene clutter that would unnecessarily obstruct the edited object motion. We only remove non-structural objects, i.e., scene elements that could plausibly be moved in a real environment, such as chairs, small tables, or other movable items, while keeping structural geometry such as walls, floors, and ceilings fixed. We distinguish structural from non-structural objects using simple heuristics based on object names and geometry; for example, walls and floors are identified either from names containing terms such as `wall', `floor', or `ceiling', or from large planar geometry. After sampling the object placement and motion, we estimate the object's expected trajectory for that motion type and sweep the object volume along that trajectory to detect scene objects it would intersect. Any intersecting non-structural objects are removed from the collision scene, after which we export the filtered collision mesh and rerun the same task with the same sampled motion parameters.

\subsection{Implementation details}
We implement data generation with Blender Cycles for rendering and Bullet for physics simulation.
Runs are executed in parallel on a GPU cluster with one render worker
per GPU.

\subsection{Dataset statistics}
\label{sec:dataset_statistics}
\nameDataset is built from curated Evermotion~\cite{evermotion_website} indoor scenes and a foreground pool
of 119 assets, comprising 98 Objaverse objects and 21 primitives with Bullet proxy variants.
After filtering, the dataset contains 33,290 paired samples, corresponding to 66,580 RGB videos and 66,580
binary mask videos.
We use 31,629 pairs for training and 1,661 pairs for testing.

Trajectory composition in the filtered set spans 18 generator-level
motion categories, which refine the high-level motion families in
Sec.~3.1 of the main paper and the procedural motion families in
Sec.~\ref{sec:object_trajectory_details}:
10,615 drag pairs, 4,778 roll pairs, 3,962 throw pairs, 3,682 drop pairs, 1,564 placement static pairs,
1,290 placement falling pairs, 783 door swing pairs, 759 pendulum pairs, 732 curved slide pairs, 720
straight slide pairs, 708 C-shaped surface path pairs, 653 falling leaf pairs, 628 8-shaped controlled-air
pairs, 533 wiggly line surface path pairs, 501 U-shaped surface path pairs, 501 stick-slip-slide pairs, 473
O-shaped controlled air pairs, and 408 Z-shaped controlled air pairs.
The 36 task variants referenced in the motion library ablation correspond to finer generator configurations
within these categories, including different path and interaction variants.

For drag templates, the counts are 3,151 O-shaped, 3,184 S-shaped, 3,202 spiral, and 1,078 point-to-point
drag pairs.
Among the 30,436 pairs for which hit/no hit annotations are defined, 13,462 are hit pairs and 16,974 are
no hit pairs.

\section{Additional Qualitative Comparisons}
\label{sec:add_qual_comparison}
\begin{figure*}[t]
    \centering
    \includegraphics[width=\linewidth]{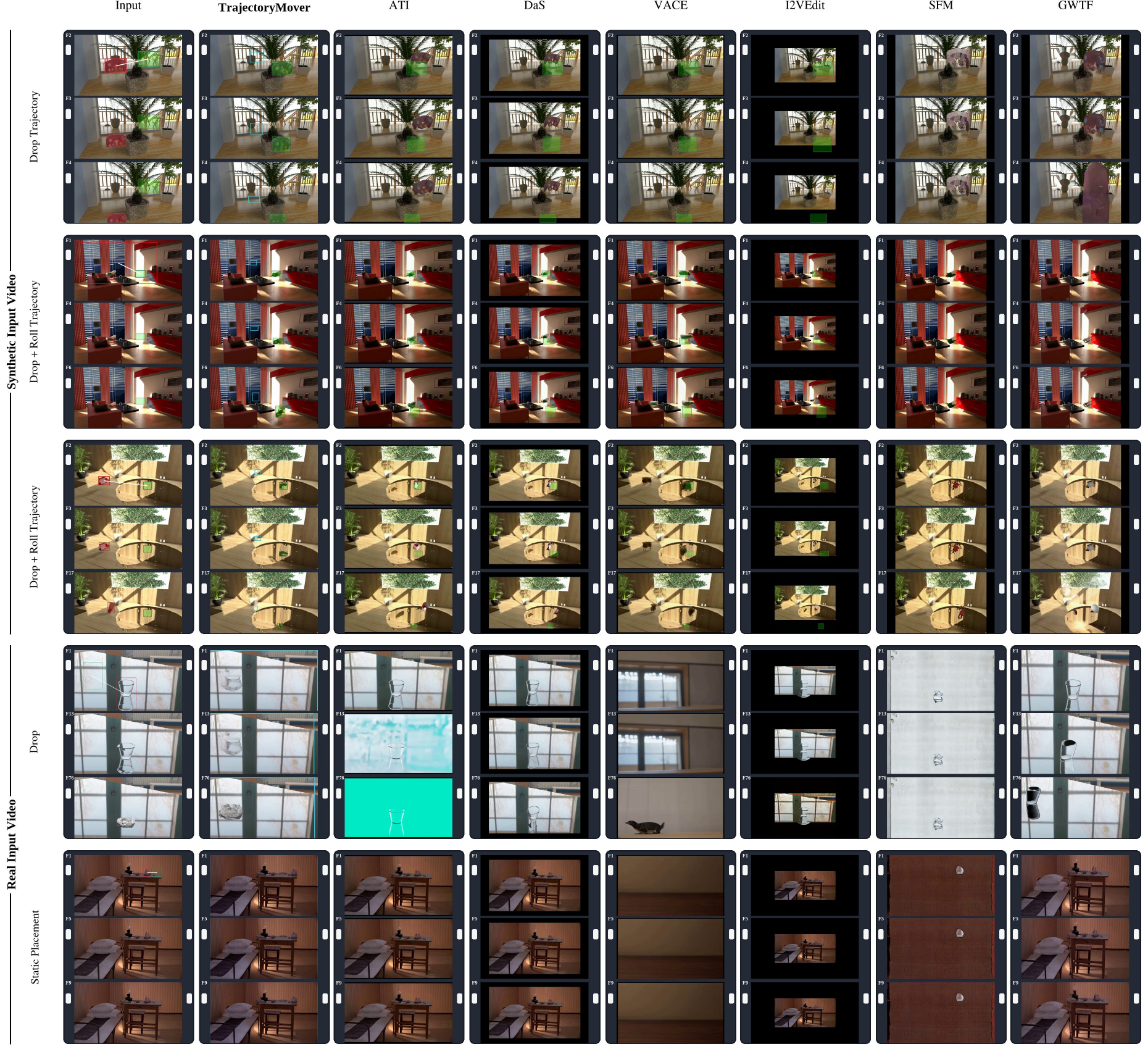}
    \caption{
        \textbf{Qualitative comparison with baselines on synthetic and real videos.}
        We compare \name{} with ATI, DaS, VACE, I2VEdit, SFM, and GWTF on representative object-relocation scenarios spanning synthetic and real videos, including drop, drop-and-roll, and static-placement motions.
        Red boxes indicate the source object location in the input video, green boxes indicate the requested target location at frame 0, cyan boxes indicate the edited object region in generated results, and white arrows indicate the intended displacement direction.
        Across both synthetic and real examples, \name{} follows the requested motion more consistently while better preserving object appearance and scene identity.
        Please zoom in for details.
    }
    \label{fig:combined_baseline_qual}
\end{figure*}

Figure~\ref{fig:combined_baseline_qual} provides additional qualitative comparisons on both synthetic and real videos.
The examples span representative object-relocation scenarios, including drop, drop-and-roll, and static-placement motions, and compare \name{} with ATI, DaS, VACE, I2VEdit, SFM, and GWTF.
Across these cases, \name{} more reliably follows the requested target placement and subsequent motion while better preserving object appearance and scene identity than the baselines.

\section{Qualitative Ablation Examples}
\label{sec:qual_ablation}
\begin{figure*}[t!]
    \centering
    \includegraphics[width=\linewidth]{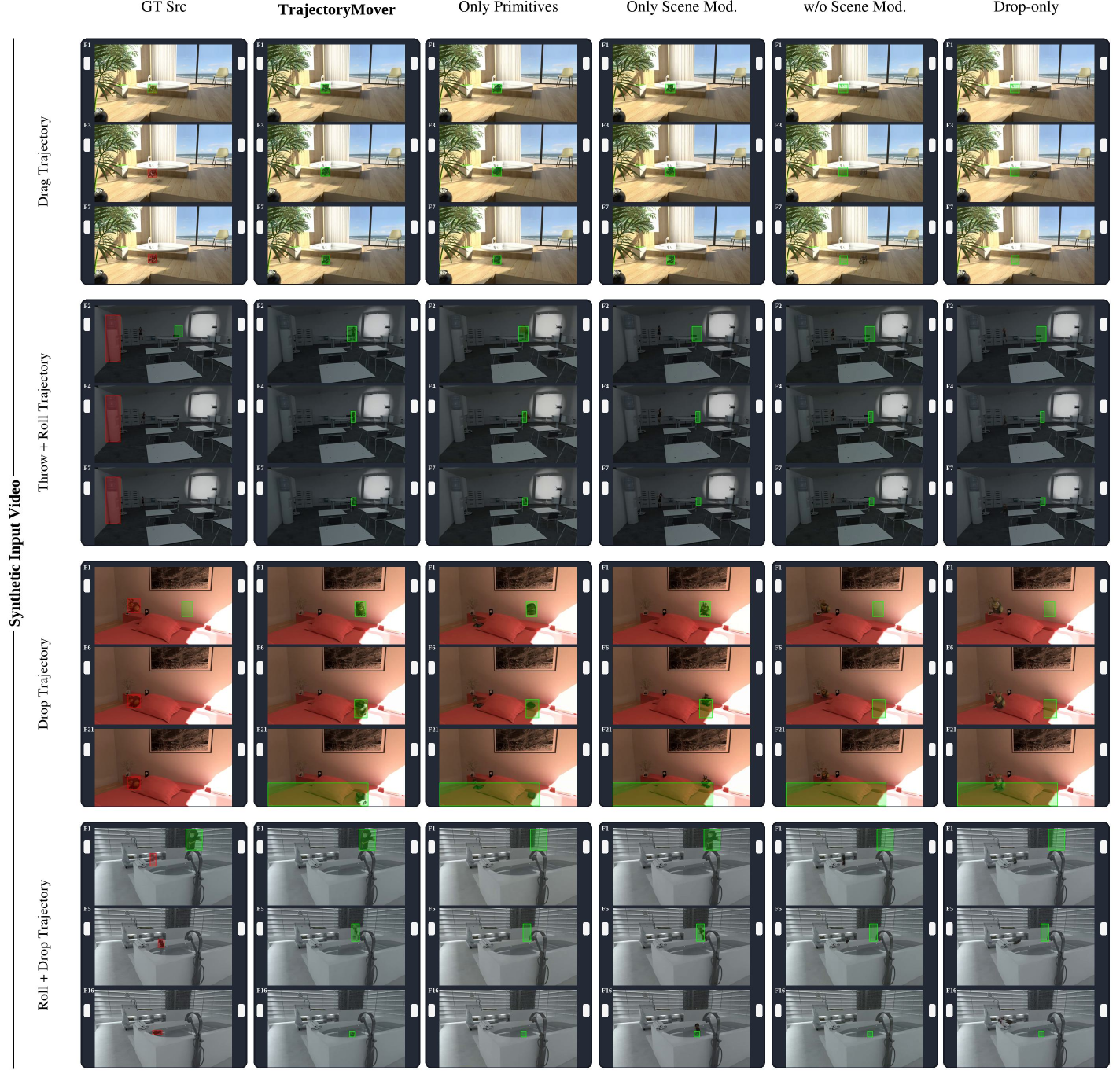}
    \caption{
        \textbf{Qualitative ablation analysis.}
        We compare the full model with ablations using only primitives, only scene modification, without scene modification, and drop-only motion training on representative synthetic motion scenarios.
        Red boxes indicate the source object location in the input video, and green boxes indicate the requested target location at frame 0.
        The full model gives the best balance of trajectory fidelity, object identity preservation, and scene-aware motion plausibility across drag, throw-and-roll, drop, and roll-and-drop trajectories.
        Please zoom in for details.
    }
    \label{fig:ablat}
\end{figure*}

We provide qualitative examples for several of the ablations in Table 3 of the main paper.
Figure~\ref{fig:ablat} shows that each data generation component contributes distinct behavior.
The full model gives the strongest combined performance in trajectory fidelity, identity preservation, and scene interaction realism.
Only primitives degrades identity most clearly, with blob-like shapes, weaker texture, and lower temporal consistency in difficult frames.
Only scene mod. helps obstacle avoidance in no-hit settings but underexposes the model to richer interactions, often reducing placement accuracy in complex scenes.
Without scene mod. improves some hit interactions but weakens learning of long-range unobstructed motion, leading to larger trajectory misalignment under changed initialization.
\texttt{Drop}-only is too narrow and transfers poorly to roll and drag motions.
Overall, the qualitative results support the full training recipe with mixed object sources, mixed hit and no-hit data, and diverse motion tasks. Please zoom in on the figure~\ref{fig:ablat} for details.

\section{User Study Interface}
\label{sec:user_study_interface}
\begin{figure}[b]
    \centering
    \includegraphics[width=\linewidth]{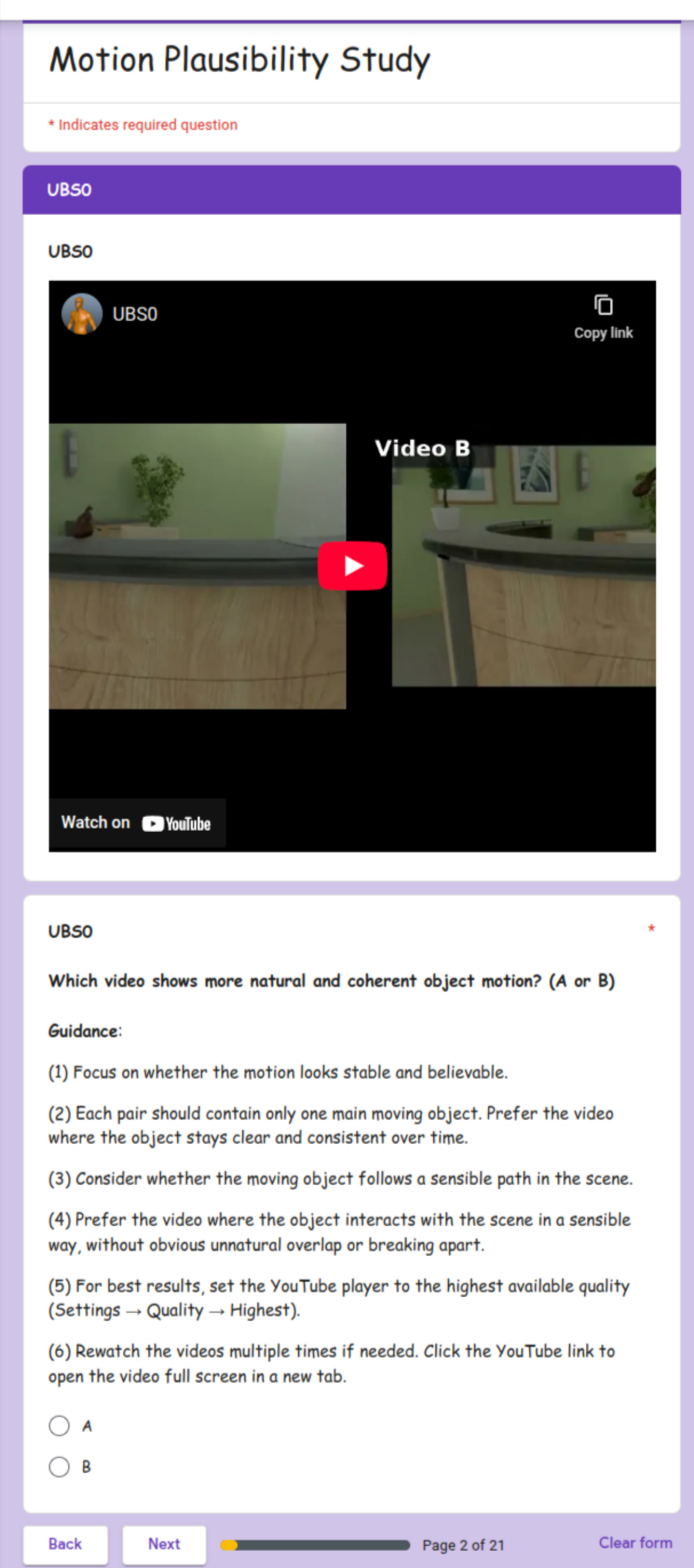}
    \caption{
    \textbf{User study interface.}
    Participants compare two anonymized videos (A/B) from the same source-target setup and select the one with more natural and coherent object motion.
    The guidance text standardizes the evaluation criteria (motion stability, object consistency, trajectory reasonableness, and scene interaction plausibility).
    The same interface format is used for all user study experiments in this project.
    }
    \label{fig:userstudy_interface}
\end{figure}

Figure~\ref{fig:userstudy_interface} shows the user interface used for the user study presented in the main paper.

\end{document}